\documentclass[sn-mathphys,Numbered]{sn-jnl}% Math and Physical Sciences Reference Style
\usepackage{graphicx}%
\usepackage{multirow}%
\usepackage{amsmath,amssymb,amsfonts}%
\usepackage{amsthm}%
\usepackage{mathrsfs}%
\usepackage[title]{appendix}%
\usepackage{xcolor}%
\usepackage{textcomp}%
\usepackage{manyfoot}%
\usepackage{booktabs}%
\usepackage{algorithm}%
\usepackage{algorithmicx}%
\usepackage{algpseudocode}%
\usepackage{listings}%
\begin{document}

\title[Article Title]{Foul prediction with estimated poses from soccer broadcast video}

\author[1]{\fnm{Jiale} \sur{Fang}}\email{fang.jiale@g.sp.m.is.nagoya-u.ac.jp}

\author[1]{\fnm{Calvin} \sur{Yeung}}\email{yeung.chikwong@g.sp.m.is.nagoya-u.ac.jp}

\author*[1,2,3]{\fnm{Keisuke} \sur{Fujii}}\email{fujii@i.nagoya-u.ac.jp}

\affil*[1]{\orgdiv{Graduate School of Informatics}, \orgname{Nagoya University}, \orgaddress{\street{Chikusa-ku}, \city{Nagoya}, \state{Aichi}, \country{Japan}}}

\affil[2]{\orgdiv{RIKEN Center for Advanced Intelligence Project}, \orgname{1-5}, \orgaddress{\street{Yamadaoka}, \city{Suita}, \state{Osaka},  \country{Japan}}}

\affil[3]{\orgdiv{PRESTO}, \orgname{Japan Science and Technology Agency}, \orgaddress{\city{Kawaguchi}, \state{Saitama},\country{Japan}}}

\abstract{
    Recent advances in computer vision have made significant progress in tracking and pose estimation of sports players. However, there have been fewer studies on behavior prediction with pose estimation in sports, in particular, the prediction of soccer fouls is challenging because of the smaller image size of each player and of difficulty in the usage of e.g., the ball and pose information. 
    % In this paper, we propose a new deep-learning method to predict soccer fouls with the combination of video, bounding box position and image and pose information by constructing a new soccer foul dataset.
    % We leverage convolutional and recurrent neural networks (CNNs and RNNs), which fuse the above four modalities. 
    In our research, we introduce an innovative deep learning approach for anticipating soccer fouls. This method integrates video data, bounding box positions, image details, and pose information by curating a novel soccer foul dataset. Our model utilizes a combination of convolutional and recurrent neural networks (CNNs and RNNs) to effectively merge information from these four modalities.
    The experimental results show that our full model outperformed the ablated models, and all of the RNN modules, bounding box position and image, and estimated pose were useful for the foul prediction.
    Our findings have important implications for a deeper understanding of foul play in soccer and provide a valuable reference for future research and practice in this area.
}

\keywords{datasets, neural networks, soccer, pose estimation}

\maketitle

\section{Introduction}\label{Introduction}

     With the rapid development in the field of computer science, computer vision has become a research field of great interest. With the continuous advancement of technology, impressive achievements have been made in the fields of object detection and tracking, and action classification, laying the foundation for subsequent analysis.
     % As computer science undergoes rapid advancements, there is a growing fascination with the research field of computer vision. The relentless progress in technology has led to remarkable accomplishments in areas such as object detection, tracking, and action classification. These achievements not only showcase the current state of the art but also establish a solid groundwork for future analyses and developments.  (言い換え)
     In sports such as soccer, larger sports video datasets (e.g.\cite{giancola2018soccernet, deliege2021soccernet,giancola2021temporally,scott2022soccertrack}) have been published and algorithms based on the datasets have been developed such as multi-object tracking (MOT) \cite{cioppa2022soccernet,scott2022soccertrack}, action spotting \cite{giancola2018soccernet,cioppa2020context,giancola2021temporally}, and video understanding \cite{giancola2023towards,held2023vars,mkhallati2023soccernet}. 
    However, most of the recent methods for action spotting or classification have directly used video as input for deep learning models \cite{giancola2018soccernet,khan2018learning,rongved2020real,karimi2021soccer}, and fewer studies have considered behavior predictions with pose information (e.g., shot prediction \cite{goka2023prediction} and pass receiver prediction \cite{honda2022pass}) because behavior prediction with pose estimation in soccer are challenging due to the smaller image size of each player and the difficulty in the usage of the ball information. 
    In the context of autonomous driving and human behavior understanding, methods of pedestrian intention prediction have been intensively investigated. 
    Earlier work used single-frame images as input into convolutional neural networks (CNNs) for the prediction \cite{rasouli2017they}, and recently spatio-temporal image sequence features such as videos, bounding boxes (bboxes), and estimated pose for the pedestrians such as using recurrent neural networks (RNNs) (e.g., \cite{piccoli2020fussi,rasouli2022multi,kotseruba2021benchmark} and reviewed by \cite{sharma2022pedestrian}). 
    
    In soccer, the prediction of foul behavior is of great importance to referees and players.
    % In soccer, predicting when a player will commit a foul is crucial for referees and players.(言い換え)
    When it comes to dangerous behaviors and violations in soccer, it has a significant impact on the outcome of the game and the safety of the players. 
    % It has a significant impact on the outcome of the game and the safety of the players when it comes to dangerous behaviour and foul play in soccer.(言い換え)
    With the help of posture analysis technology, we can expect to predict the presence of foul play by analyzing a player's posture and movements. For example, whether a player has made a deliberate foul move or inappropriate physical contact, potential foul play can be predicted in advance and provides an objective basis for decision-making. 
    % Through the utilisation of posture analysis technology, there exists the potential to proactively predict the manifestation of foul play within the domain of soccer. This is achieved by methodically scrutinising the nuanced postures and kinetic patterns exhibited by players. For instance, the deliberate execution of a foul manoeuvre or engagement in improper physical contact can be discerned through meticulous analysis. Consequently, the anticipation of potential foul play provides an anticipatory framework, supplying an empirically grounded basis for judicious decision-making processes within the realm of sports officiation.(言い換え)
    However, challenges still exist in predicting foul plays in soccer as described above.

\begin{figure}[h]
  \centering
  \includegraphics[width=\linewidth]{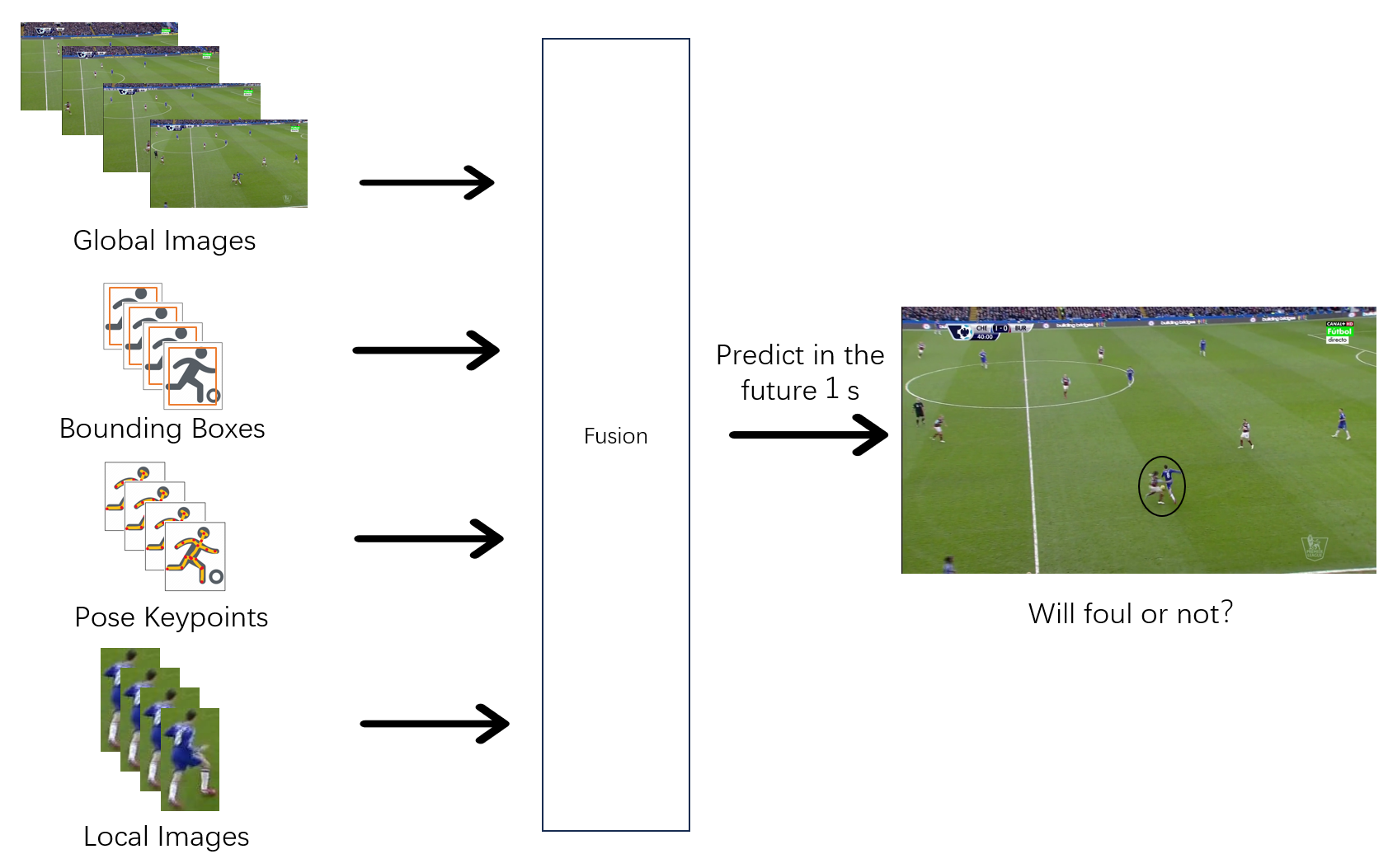}
  \caption{Our foul prediction system (FutureFoul). Our method uses video, bbox, bbox image and pose information of 3 s duration to predict fouls in the future 1 s.}
  \label{fig:picMain1_new}
\end{figure}

    % proposed method
    In this paper, we propose a foul behavior prediction system called \textit{FutureFoul} from soccer broadcast videos, as shown in Figure \ref{fig:picMain1_new}. The system analyzes data such as the player's posture, movement, and position and compares and evaluates them.
    In the experiments, we show that our system can accurately identify potential fouls and provide appropriate warnings or alerts. 
    We also show that in a soccer game, players' postures and movements are crucial for determining the presence of foul play. 
    By utilizing our approach, we hope to help the management and supervision of soccer games, facilitating the development of the game and improving the safety of players.

    The contributions of this paper are as follows: 
    \begin{enumerate}
      \item We propose a new deep learning method to predict soccer fouls with the combination of video, bounding box position and image, and pose information by constructing a new soccer foul dataset. We leverage the fused CNNs and RNNs for an accurate foul prediction. 

      \item The experimental results show that our full model outperformed the ablated models, and the RNN module, bounding box position and image, and estimated pose were useful for the foul prediction. We also analyzed the successes and failures in the predictions and obtained insights about the foul prediction.
    \end{enumerate}
    
\section{Related work}\label{Related work}
% 1st paragraph
\subsection{Multi-object tracking and datasets in soccer}
    As a large broadcast video dataset in soccer, the SoccerNet \cite{giancola2018soccernet} dataset provides a resource of recorded video of soccer leagues in Europe, but the dataset lacks detailed information such as annotations, which makes it relatively difficult to use. To overcome this problem, the SoccerNet-v2 \cite{deliege2021soccernet} dataset adds about 300,000 annotations to SoccerNet, extends the range of research tasks in the field of soccer, including action spotting, camera shot segmentation with boundary detection, and defines a novel replay-based task.
    In addition, the SoccerNet-v3 \cite{cioppa2022scaling} dataset annotates action replay sequences from soccer matches with annotations that include lines and goal sections on replay frames and live action frames, as well as annotating bboxes for players on the field and specifying the team to which the player belongs. 
    For other similar datasets, SportsMOT \cite{cui2023sportsmot} and SoccerTrack \cite{scott2022soccertrack} have been published.

    %About character recognition and tracking, 
    In MOT, various methods have been developed mainly for pedestrian tracking, mainly with tracking-by-detection paradigms such as DeepSORT \cite{veeramani2018deepsort} and ByteTrack \cite{zhang2022bytetrack}. 
    Recently, end-to-end methods utilizing Transformer have been developed such as \cite{wang2021end,carion2020end}.
    In soccer MOT, the SoccerNet Tracking \cite{cioppa2022soccernet} tested several tracking methods including DeepSORT \cite{veeramani2018deepsort}, FairMOT \cite{zhang2021fairmot}, and ByteTrack \cite{zhang2022bytetrack}. % The authors fine-tuned 
    % The proposed large-scale sports scene multi-target tracking dataset of 
    SportsMOT \cite{cui2023sportsmot} proposed MixSort, a new multi-object tracking framework, with introducing a MixFormer-like structure \cite{cui2022mixformer}.
    In re-identification, the recent approach Body Feature Alignment Based on Pose \cite{akan2023reidentifying} successfully extracts key features of the players in the image, overcoming challenges such as the same team wearing similar outfits, limited samples, and low-resolution images.
    A soccer and player tracking method based on You Only Look Once (YOLOv3) and Simple Online Real-Time (SORT) \cite{naik2023yolov3} was also proposed, aiming to accurately categorize objects detected in soccer videos and track them in a variety of challenging contexts.
    A semi-supervised training method \cite{vandeghen2022semi} also  
    % was introduced to enhance player and ball detection in soccer. The method 
    utilized a combination of labeled and unlabeled data to improve the performance of the detection algorithms.
     In this study, we used ByteTrack \cite{zhang2022bytetrack}, which uses complex association and matching algorithms to achieve accurate MOT, because our aim is not to improve the MOT. % We can use Bytetrack \cite{zhang2022bytetrack} to improve the accurate tracking ability of our player pose analysis and soccer foul behavior prediction models.

     % 2nd paragraph: 
     \subsection{Behavior recognition and prediction in soccer}
    % \noindent \textbf{Behavior recognition and prediction in soccer.}
     Since there have been many studies of behavior recognition in sports, we introduce soccer-related work. 
    As behavior recognition, for instance, action spotting \cite{giancola2018soccernet,cioppa2020context,giancola2021temporally,deliege2021soccernet} have been recently investigated. As an example, a temporally-aware feature pooling \cite{giancola2021temporally} approach was proposed for action spotting in soccer broadcasts. 

    Another research direction is foul detection, a joint architecture \cite{khan2018learning} is proposed for detecting foul events, which combines event clips with contextual information and enables the localization of events and proposal of story boundaries.
    % ,C3D \cite{khan2018learning}, 
    % \cite{thamaraimanalan2020prediction}
    %
    In general, OpenPose \cite{cao2017realtime} , HRNet \cite{sun2019deep}, OCHuman \cite{zhang2019pose2seg}, HigherHRNet \cite{cheng2020higherhrnet}, BalanceHRNet \cite{li2023balancehrnet} and more have been used for pose estimation.

    Although the behavior prediction has been investigated mainly from event and location data (e.g., \cite{decroos2019actions,toda2022evaluation,simpson2022seq2event,yeung2023transformer,umemoto2022location}), % VAEP, VDEP, Seq2Events, NMSTPP, GVDEP
    there have been few prediction methods from video (e.g., shot prediction \cite{goka2023prediction} and pass receiver prediction \cite{honda2022pass}). Video footage of a soccer game captures the postures of the players, offering valuable insights into their future actions and intentions. 
    Therefore, in this study, we utilize the video footage to predict the foul with estimated pose information by OCHuman \cite{zhang2019pose2seg}, which is challenging due to the smaller image size of each player and the difficulty in the usage of the ball information.

\section{Dataset}\label{Dataset}

    We construct the soccer foul dataset to verify our methods. In this section, we first describe the video dataset we used, and then explain the creation of foul labels, bboxes, and pose estimation. 

\subsection{Video dataset}\label{Video dataset}

    The videos are obtained from SoccerNet-v3 dataset \cite{cioppa2022scaling}, which is a code base for a generic data loader containing annotations and data in SoccerNet Dataset \cite{giancola2018soccernet}. 
    The SoccerNet dataset is a large-scale dataset for soccer video understanding. It contains 550 complete broadcast soccer matches and 12 single-camera matches from major European leagues. The video data in the SoccerNet dataset is provided in .mkv format with a frame rate of 25 frames per second and a resolution of 720p or 224p. 
    SoccerNet-v3 provides a convenient way to load images and parse JSON annotation files for correspondence between bboxes, lines, and bboxes, thus providing a rich environment for soccer-related computer vision tasks. Multiple replay sequences of 500 fully broadcast matches were covered in the SoccerNet-v3 dataset. For each action in the replay sequence, the timestamps of the replay frames showing the real-time action frames were marked and these replay frames and action frames with lines and goals were annotated. In addition, each character on the field with a bboxes and specified the team they belonged to were annotated. Also, prominent objects were annotated. To establish player correspondence between replay and live view, jersey numbers were used as identifiers whenever possible. Such data annotation makes the SoccerNet-v3 dataset ideal for processing soccer-related computer vision tasks.

\subsection{Selection of foul labels}

    First, we selected the foul tag ``foul'' from the Soccernet-v3 dataset \cite{cioppa2022scaling}. In addition, we also selected other tags for the non-foul tags, including ``Ball out of play'', ``Clearance'', ``Shots on target'', ``Shots off target'', ``Offside'', and ``Goal''. These tags provide references to different scenarios and actions in the game and are used to build a comprehensive dataset. It is worth noting that some of the tags were not used in this experiment due to excessive camera movement and long ball movement distances.
    In addition, in this study, we focus on the prevention of fouls arising from dangerous behavior. Therefore, we decided not to consider the ``offside'' in soccer as a foul. Offside is an offense under the rules of soccer when an attacking player receives the ball behind the opposing defense without at least two defending players in front of him. Although an offside infringement can affect the play and outcome of a match, it is not an infringement that directly involves dangerous behavior. Our research aims to explore and prevent behaviors that could lead to injury or danger. Therefore, we have chosen to focus on those acts that may trigger direct physical contact, foul play, or potential injury. To make it easier to identify players who committed fouls, we added a screening criterion. We excluded data on fouls for which the soccer could not be found in the image. Finally, we successfully got data for 2,500 fouls and 2,500 non-fouls.

\begin{figure}[h]
  \centering
  \includegraphics[width=\linewidth]{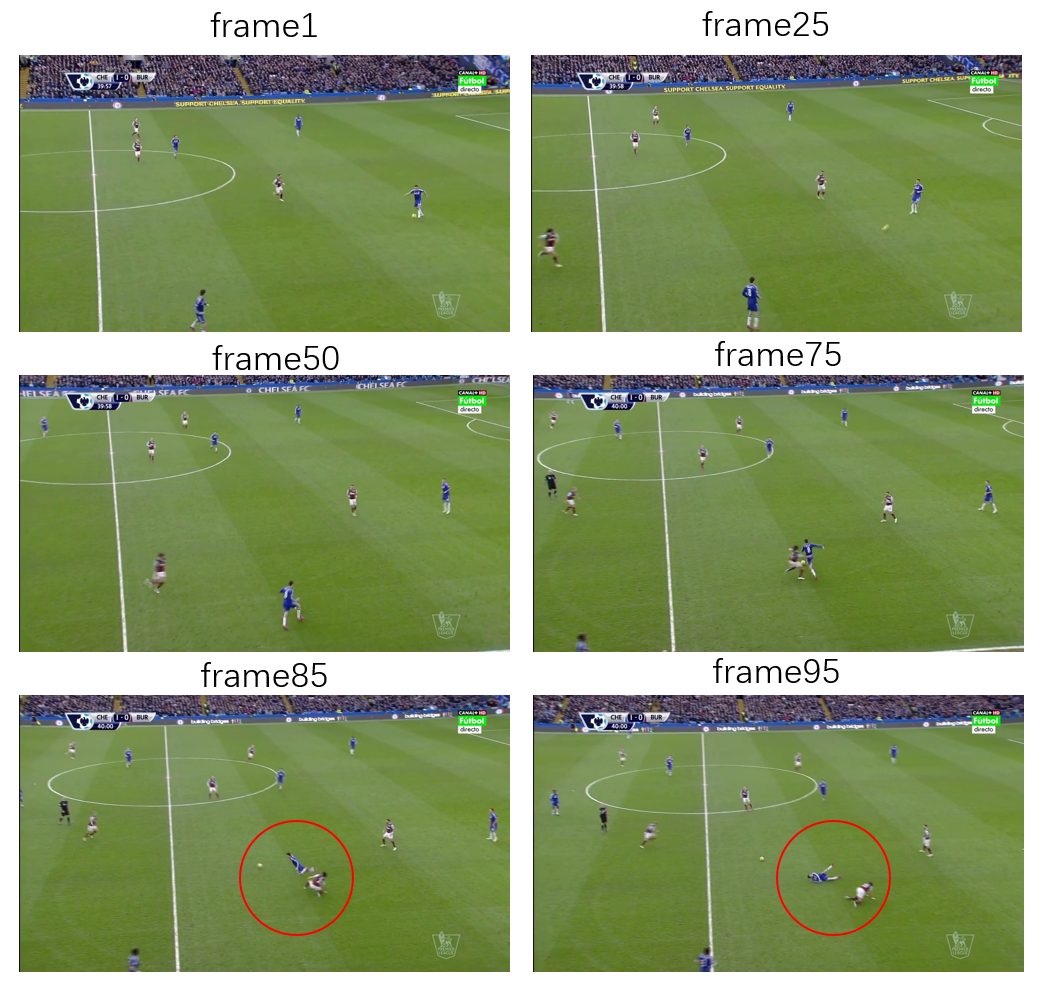}
  \caption{Video data example. Frames 1, 25, 50, and 75 are the time-ordered frames in the 3 s before the foul. Frame 85 and 95 are examples of foal happening within 1 s of the denoted foul time, which is not be used for our FutureFoul model training.}
  \label{fig:picVideo_new_withmark}
\end{figure}

\subsection{Extraction of analysis interval}

    Here we explain the extraction of the analysis interval in the video dataset. 
    First, we filter the events labeled "foul" from Soccernet-v3 dataset \cite{cioppa2022scaling} and record the number of matches and the exact time of these events. We then use the video footage from the Soccernet dataset to cut out clips related to these events. Our cropping method is to select video from 3 seconds before and 1 second after the denoted foul time, for a combined duration of 4 seconds. Such a selection is based on the minimum granularity of the event occurrence time in seconds in the Soccernet-v3 dataset. Considering that the time stamp in the dataset may be inaccurate by up to 1 second (for example, although the foul happens at 3 min 30s 12ms, the time step from Soccernet-v3 will denote as 3min 30s), we ensured that the cropped video covered the key moments before and after the actual foul event. Through this process, we are able to obtain data with the ``foul'' tag and extract the relevant game, the specific time, and the corresponding video clip. This data and video footage will support our research and allow us to gain insight into the characteristics and patterns of foul play. An example of a foul event video data is shown in Figure \ref{fig:picVideo_new_withmark}. We took the first 75 frames of each video as training data.

\begin{figure}[h]
  \centering
  \includegraphics[width=\linewidth]{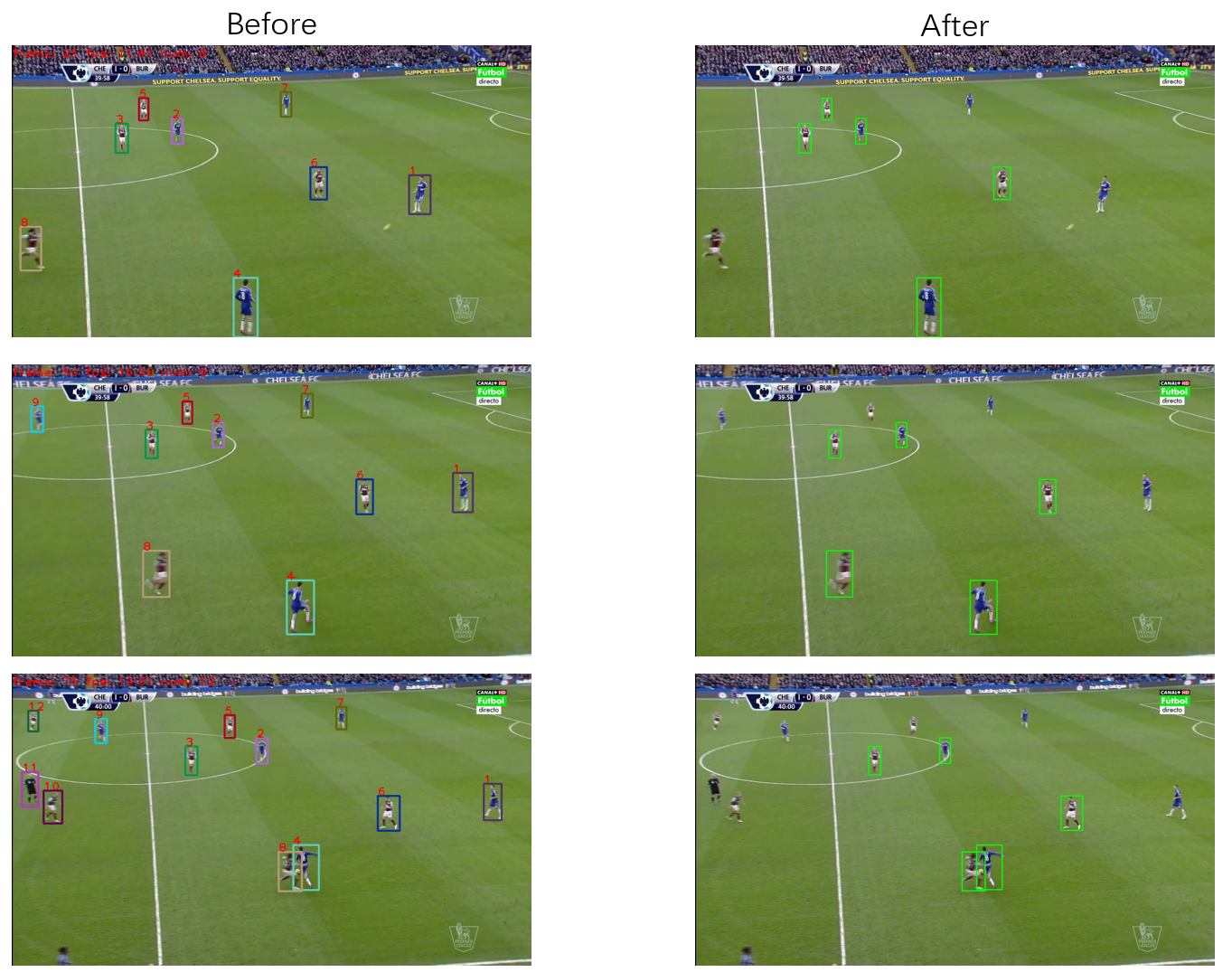}
  \caption{Bbox extraction. The bbox information extracted using ByteTrack \cite{zhang2022bytetrack} (Before) was filtered to extract the five closest to the position of the soccer ball at the time of the foul (After).}
   \label{fig:picVideo_bbox}
\end{figure}

\subsection{Extraction of bboxes with object tracking}
To extract the bbox data from the video, we used the existing ByteTrack model \cite{zhang2022bytetrack}. However, as there are different numbers of people in each frame, some decisions were made to facilitate model learning. Typically, at the time of a foul event, there may be 2-3 people who appear to be falling. However, it is challenging to accurately locate these 2-3 people. To solve this problem, we decided to choose a fixed number of people as the observation target and set it to 5. We took the frame where the foul occurred and selected the five people closest to the ball on that frame as our observation target. By performing a backtracking approach,
We identify the five players closest to the soccer ball in the foul frame and their respective bboxes. For the previous frame, we track the 5 players by finding the closest bboxes with the center point of the bbox. In this step, we also manually checked the data and removed some inaccurate data.
The purpose of this approach is to ensure that we obtain the key person movement data and to reduce the complexity of the model. By fixing the number of people, we can better capture the relevant movement trajectories and provide a more accurate basis for data analysis and research. An example of a foul event bbox data is shown in Figure \ref{fig:picVideo_bbox}.

\begin{figure}[h]
  \centering
   \includegraphics[width=\linewidth]{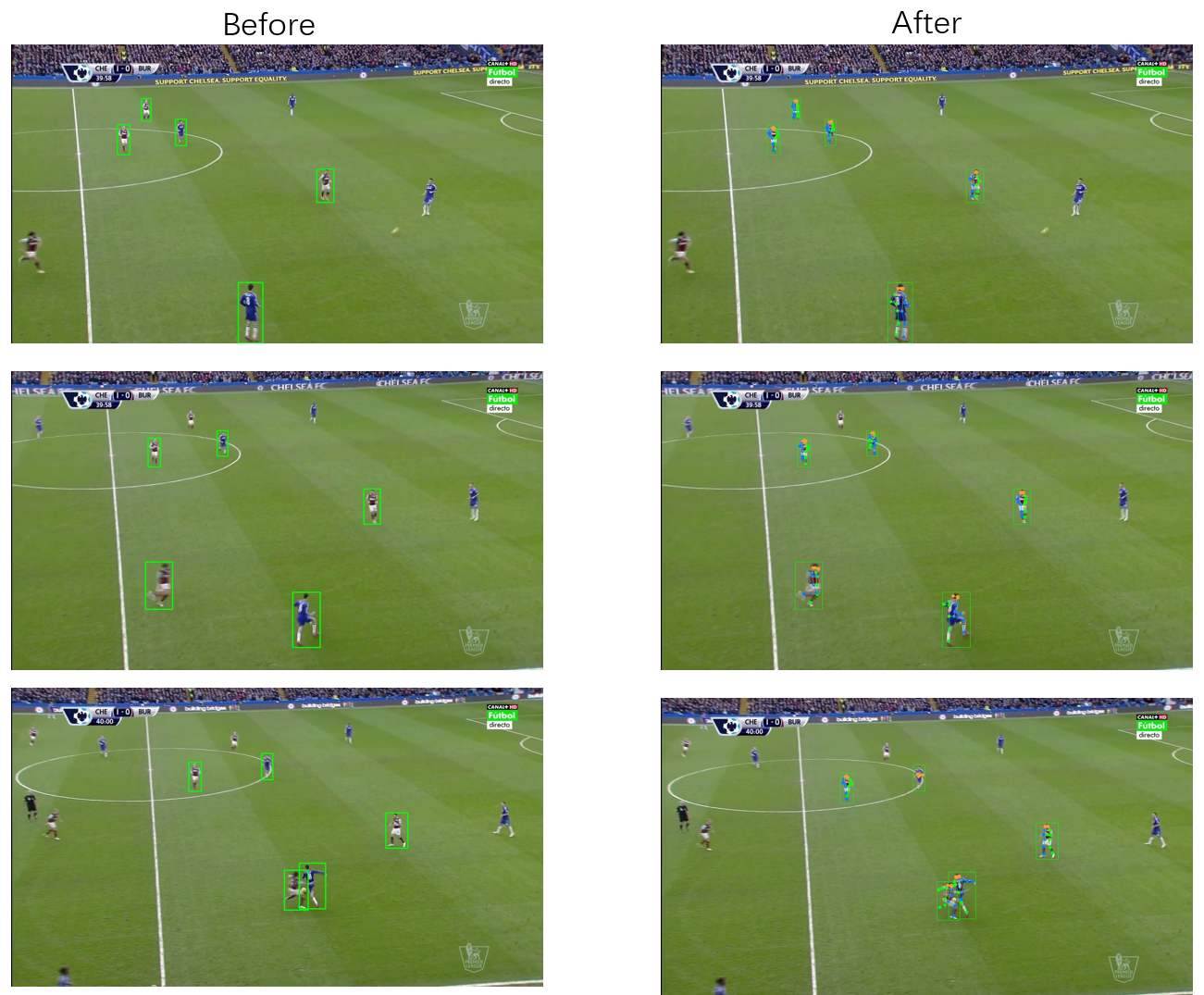}
  \caption{Pose extraction. We used OCHuman \cite{zhang2019pose2seg} to obtain pose data (After) corresponding to the previously obtained Bbox information (Before).}
  \label{fig:picVideo_pose}
\end{figure}
\subsection{Extraction of estimated pose information}
In the step of extracting pose information, we make use of the previously acquired bbox data and video data. This data is fed into a pre-trained ResNet-50 network and processed using an OCHuman model \cite{zhang2019pose2seg}. This OCHuman model was pre-trained based on the ResNet-50 network. With the OCHuman model, we were able to extract information about the poses of five people at different points in time, including 17 key points. These key points are nose, left eye, right eye, left ear, right ear, left shoulder, right shoulder, left elbow, right elbow, left wrist, right wrist, left hip, and right hip. These key points provide important information about human posture. They are crucial to the analysis and understanding of human movement and the characteristics of posture. With the OCHuman model, we were able to extract the posture information of these five people at various points in time from the video, providing a wealth of data for subsequent analysis and research. This data can be used for tasks such as action recognition, pose analysis, and human behavior pattern recognition, providing us with strong support for the in-depth study of action and pose in soccer matches. Figure \ref{fig:picVideo_pose} is an example of the pose data we obtained.

\begin{figure}[h]
  \centering
  \includegraphics[width=\linewidth]{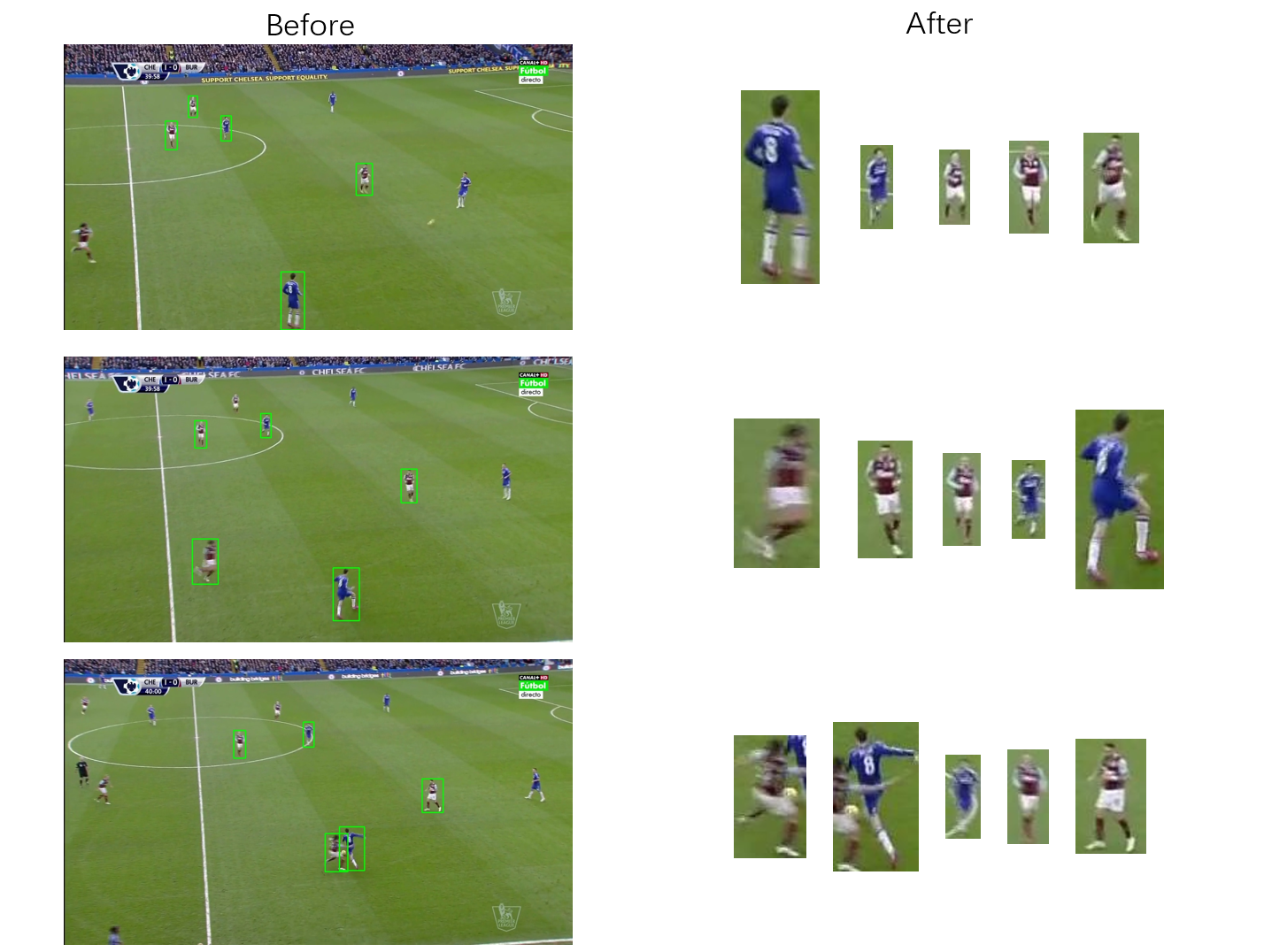}
  \caption{Bounding box image extraction. We obtain bbox image data (After) from the video data(Before).}
  \label{fig:bboxpic}
\end{figure}

\subsection{Extraction of bbox image}
In the step of extracting bbox picture information, we use the acquired bbox data and video data. We use bbox positions data to cut out images of the corresponding players from the video. Figure \ref{fig:bboxpic} is an example of the bbox image data we obtained.

\section{Proposed method}

    In this study, a video-based method that combines pose information and bboxes information is proposed for predicting foul behaviors in soccer games. The method is divided into the steps of feature extraction, CNN and gated recurrent unit (GRU) as RNN network modeling, and foul behavior prediction as presented in Figure \ref{fig:bboxpic}.

\begin{figure}[h]
  \centering
  \includegraphics[width=\linewidth]{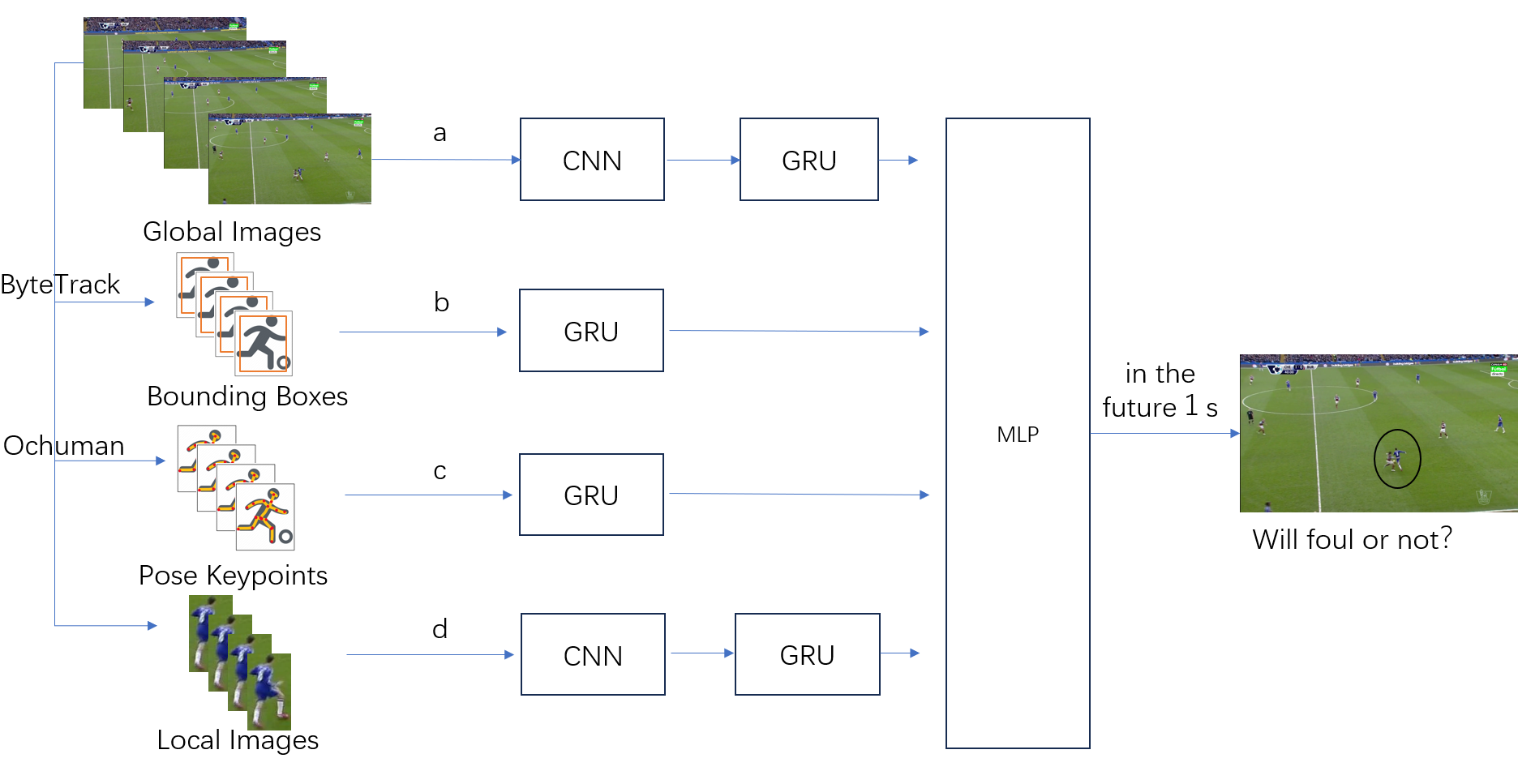}
  \caption{FutureFoul model architecture. We construct a four-branch network structure for foul behavior prediction. (a),(d) Vision Branch extracts features from the four images in the video, by feeding these features into a CNN and followed by a GRU network the GRU last time step output is obtained. (b), (c) Bbox position and Pose Branches have a similar structure, they input the bbox information and pose information into GRU networks respectively.
  Finally, by concatenating all four benches' output and feeding into an MLP, we predict whether a foul will be committed in the future 1 s or not.}
  \label{fig:picMain_new}
\end{figure}

\subsection{Feature creation}
    In order to create features, we need to pre-process the data obtained above to ensure the quality and consistency of the data. As an input of our model, we subsampled video frames such as four frames from each video, frame 1, frame 25, frame 50, and frame 75 (we examined various subsample frequencies in the experiments), and obtained the bbox and pose data on these frames. For memory saving and faster training, in pass receiver prediction in Soccer Videos \cite{honda2022pass}, they down-sampled the videos. According to this, we also down-sampled the video to speed up our training. We will further investigate the effect of skipping in future studies. We created the following four types of features. 

    \begin{enumerate}
     \item Video: To facilitate the learning of the network, We resized the input image to a size of $64 \times 64$ (in the experiment, we validate the sizes). The purpose of resizing the images is to meet the input requirements of the network model. The smaller image size helps to reduce the number of parameters and computational complexity of the model, improving the efficiency of training and inference.
      \item Bbox: We are particularly interested in the position information and movement information of the players. To do this, we extracted the foot position (the bottom center of the bbox) of each player at each frame from the bbox data and used this as input to the model. We did not use the soccer ball's bbox data during training, which was only used for preprocessing.
      \item Pose: The key points in the pose data may not be detected completely due to occlusion. To deal with this situation, we take a padding operation and fill the undetected parts with zeros in order to facilitate model learning.
      \item BboxImg: We want to focus more on the actions of athletes to predict fouls more accurately, so we extracted images of five athletes from the video and use bbox's image data as an input to the model.

    \end{enumerate}

    In this way, we were able to provide the model with accurate training and testing data and enable it to effectively learn and understand information about players' positions, movements, and postures. These steps could help to improve the performance of the model and its ability to recognize different events in the game.

\subsection{Proposed model: FutureFoul}
    To model the features temporally, we use the RNN (GRU) model. The RNN model architecture is capable of capturing athlete action sequences and temporal information. The modeling is performed by taking the pose, bbox, bbox image and video feature sequences of each frame as input, after multiple layers of RNN units.
    
    The overall architecture is shown in Figure \ref{fig:picMain_new}. It consists of CNN modules, RNN modules, and feed forward modules.

    \noindent {\bf{CNN module.}} The input to the model is a 3D array of size [3,224, 224] for image RGB channels, pixel width, and pixel height, respectively. The model processes this input data using three convolutional layers, each followed by a ReLU activation function and concludes with a maximum pooling layer. The convolutional layers utilize three 3x3 kernels and determine the number of output channels to capture distinct features. The batch normalization layer is employed to accelerate training and enhance the generalization performance of the model, while the Dropout layer helps mitigate the risk of overfitting.

    \noindent {\bf{RNN module.}} After the CNN part we use a recurrent neural network (RNN) specifically a GRU. The GRU receives the features output from the CNN part, the bbox, or pose keypoints and processes the features as a time series. the GRU layer has 2 hidden layers with a hidden state dimension of 256. the role of the GRU is to capture the temporal dependencies by memorizing and updating the hidden states. In our model, the last time step output of the GRU is passed to the subsequent linear for processing and prediction.

    \noindent {\bf{MLP module.}}
    MLP (Multilayer Perceptron), is a common feed-forward neural network structure. It consists of multiple hidden layers, where each neuron is connected to all neurons in the previous layer. 
    Taking these videos, bbox positions, bbox images and pose data as inputs, MLP learns to extract patterns of foul behavior from these features. It can map the input data to label for representing situations whether foul play and non-foul play situations will occur.
    
\subsection{Training}

    In the model training phase, we train the GRU networks using the labeled foul behavior data. 
    We input the bbox data, the pose data, and the picture data into their respective branches. At the same time we use bbox data and video data to extract images of athletes and input into the bbox image branch. The results obtained from each branch are then integrated into the feed forward mechanism, resulting in a prediction of whether a foul will be committed next. To train the model, we chose a learning rate of 0.001, and a batch size of 32. We used the Adam optimizer to update the model parameters. To prevent overfitting, we adopted the Dropout layer, which randomly discards some neurons during the training process to reduce the overfitting of the model to the training data. During the training process, the main metrics we focus on are training loss, validation loss, training accuracy, and validation accuracy. We expect these metrics to decrease gradually as the training progresses to ensure that both the training and validation performance of the model is improved.
    
    For the loss function, we have chosen Cross Entropy Loss, a loss function for classification problems. In our task, even though we are doing a prediction task, it is essentially still a binary classification task, where we classify the first 3 seconds of a play that will be followed by a foul player in the following second or not.

\section{Experiments}
In this section, we performed the experiments to verify our models and qualitatively analyze the successes and failures of the prediction and discuss the possibility of practical usage.

\subsection{Setup}
We used our dataset for verification of our methods, both of which are described above. 
To train our model, we first split our dataset into 4,000, 500, and 500 samples as training/validation/test sets. 
Here we compared our methods to four baseline methods with our full model: 

    \begin{itemize}
   
    \item {\verb|CNN w/ video|}: Using only video data we trained only CNN models.
    
    \item {\verb|CNN+GRU w/ video|}: Using video data as input, but we trained CNN and GRU models.
    
    \item {\verb|CNN+GRU w/ video+bbox|}: Using a combination of video and bbox position data as input, we trained the CNN and GRU models.

    \item {\verb|CNN+GRU w/ video+bbox+pose (FutureFoul w/o BboxImg)|}:  Using a combination of video, bbox position, and pose data as input, we trained the CNN and GRU models.
    
    \item {\verb|FutureFoul (CNN+GRU w/ video+bbox+pose+BboxImg)|}: This is our full FutureFoul model. Using a combination of video, bbox position, pose and bbox image data as input, we trained the CNN and GRU models.

    \end{itemize}

\subsection{Quantitative evaluation}

In this subsection, we investigated the accuracy of our models. 
Table \ref{tab:acc} shows a comparison of the accuracy of our different models. Our model (FutureFoul) includes CNN+GRU models using video, bbox, pose information and bbox image.
% , as well as other models using different combinations of inputs. 
As can be observed from the table, the FutureFoul model achieved the highest accuracy in our experiments, with 77.4\%. %This indicates that models combining video, bounding box and pose information are more accurate when predicting foul behavior.
Compared with other models, the accuracy of the CNN+GRU model using video and bbox position and pose was 74.8\%, the accuracy of the CNN+GRU model using only video and bbox position was 71.2\%, which was slightly higher than the accuracy of the CNN+GRU model using only video (67.6\%). In contrast, the accuracy of the video-only CNN model was the lowest at 65.4\%. This suggests that the RNN module, bbox position, bbox image and pose information play a role in improving the accuracy of the model in the foul prediction task.

The tendency of the results in terms of precision was similar for all models. However, results of the recall were relatively lower than that of the precision, and our full model (FuturePose) did not improve from baselines in terms of recall. 

In general, by combining pose information and bbox information, and integrating RNN networks for modeling and prediction, our method was able to accurately predict the occurrence of foul behavior in soccer games. This method provided valuable decision support for referees and coaches and could improve the fairness and quality of the game.

\begin{table}[h]
  \caption{Performances of our models. Accuracy (Acc), precision (Prec), and recall (Rec) were computed.}
  \label{tab:acc}
  \begin{tabular}{cccl}
    \toprule
    Model &Acc (\%) & Prec (\%) & Rec (\%) \\
    \midrule
     FutureFoul (ours) & \bf{77.4} & \bf{79.0} &  75.0 \\
     CNN+GRU w/ video+bbox+pose  & 74.8 & 76.0 &  72.0 \\
     CNN+GRU w/ video+bbox & 71.2 & 73.0 & 67.0 \\
     CNN+GRU w/ video & 67.6 & 63.0 & \bf{84.0} \\
     CNN w/ video & 65.4 & \bf{79.0} & 42.0 \\
  \bottomrule
\end{tabular}
\end{table}

To validate the numbers of frames to input the model, we experimented with different frame settings, including 8, 15, 25, 35, and 45 frames (Table \ref{tab:frame}). However, we did not observe improvements in accuracy with an increase in the number of frames. There was even a slight decrease.

Note that increasing the number of frames extends the training time. Given this consideration, we ultimately opted for a configuration with 4 frames. We believe it strikes a balance between accuracy and training efficiency in practical terms.

\begin{table}[h]
  \caption{Performances of our models in different frames. Accuracy (Acc), precision (Prec), and recall (Rec) were computed.}
  \label{tab:frame}
  \begin{tabular}{cccl}
    \toprule
    Model &Acc (\%) & Prec (\%) & Rec (\%) \\
    \midrule
     FutureFoul (4 frames,ours) & \bf{77.4} & \bf{79.0} &  75.0 \\
     FutureFoul (45 frames)  & 75.8 & 75.0 &  77.0 \\
     FutureFoul (35 frames)  & 73.6 & 72.0 &  77.0 \\
     FutureFoul (25 frames)  & 74.0 & 74.0 &  74.0 \\
     FutureFoul (15 frames)  & 77.0 & 77.0 &  77.0 \\
     FutureFoul (8 frames)  & 76.0 & 75.0 &  \bf{78.0} \\
  \bottomrule
\end{tabular}
\end{table}

To validate the global content image sizes, we conducted experiments with the final model with different global content image sizes, including $128 \times 128$ and $256 \times 256$ (Table \ref{tab:size}). We found that different sizes of global content images had different effects on both the training and inference phases of the model.

First, using a global content image size of $64 \times 64$, we observe that the model exhibits high accuracy (77.4\%) and precision (79.0\%) at this setting. This means that the model has a better overall performance for classification and recognition tasks of images. Meanwhile, the recall rate is 75.0\%, which indicates that the model successfully captures most of the positive case samples, but there may be some missed recognition in some cases.

Using a global content image size of $128 \times 128$, we found that the accuracy and precision of the model slightly decreased at this size, 73.2\% and 76.0\%, respectively. This may indicate that the increase in image resolution has some degree of negative impact on model performance. The recall is 68.0\%, showing a decrease in the model's performance in capturing positive examples.

Finally, using a global content image size of $256 \times 256$, we noticed a further decrease in model performance. The accuracy is 66.8\%, precision is 75.0\%, and recall is only 50.0\%. This suggests that larger image sizes may cause the model to struggle to capture all positive examples and there may be information overload

\begin{table}[h]
  \caption{Performances of our models in different size. Accuracy (Acc), precision (Prec), and recall (Rec) were computed.}
  \label{tab:size}
  \begin{tabular}{cccl}
    \toprule
    Model &Acc (\%) & Prec (\%) & Rec (\%) \\
    \midrule
     FutureFoul ($64 \times 64$,ours) & \bf{77.4} & \bf{79.0} &  75.0 \\
     FutureFoul ($128 \times 128$)  & 73.2 & 76.0 &  68.0 \\
     FutureFoul ($256 \times 256$)  & 66.8 & 75.0 &  50.0 \\
  \bottomrule
\end{tabular}
\end{table}

\subsection{Qualitative evaluation}
    To analyze the prediction results in our approach, we qualitatively analyze the example results of successes and failures from the tests.

    \subsubsection{Examples of success predictions} 
\begin{figure*}[h]
      \centering
      \includegraphics[width=\linewidth]{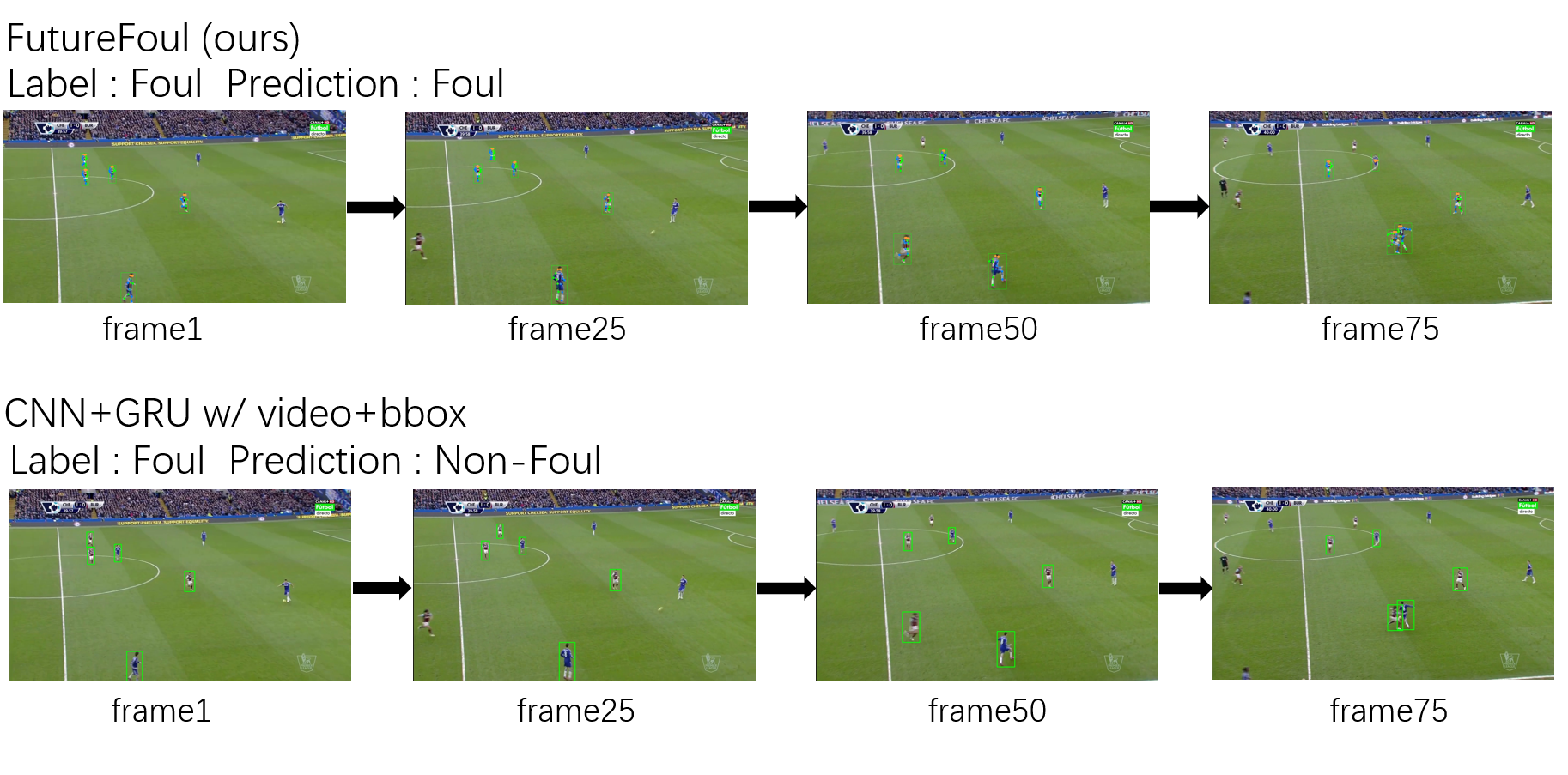}
      \caption{Qualitative evaluation using examples of success prediction. Under the same conditions, using our full model (top), with the addition of pose data to the baseline model (bottom), the foul can be successfully predicted. In contrast, CNN+GRU w/video+bbox could not predict an impending foul. }
      \label{fig:success1}
    \end{figure*}

In most cases, the model was able to accurately predict whether a foul was about to occur and gave the correct prediction. Figure \ref{fig:success1} Top, which is our full FutureFoul model, demonstrate that the model could capture key information in the video and identify patterns associated with imminent fouls.
    In Figure \ref{fig:success1} Bottom, which was CNN+GRU w/ video+bbox, we obtained limited information when we only considered the position and size of the bbox (without pose). In frame 75, the top two people were in the bbox, which could have included the impending fouler. However, if we did not consider pose information, the model could only observe that they partially overlap and was unable to determine this as an impending foul.
    However, when we added the pose key points in CNN+GRU w/ video+bbox in Figure \ref{fig:success1} Bottom, which consisted of the magnitude of the foot movement of the two people. With these pose key points, the model was able to more accurately determine that they were about to commit a foul as Table \ref{tab:acc} shows.

    \subsubsection{An examples of failures: missed detection}

    The system sometimes incorrectly predicted the fouls as non-foul (Recall: 75.0\%). Although we expected the model to accurately identify and predict potential foul play, in some complex scenarios, the model could fail to capture all the details and features, resulting in a failed prediction. 
    This could have happened because there were many variables and uncertainties in the scenes, and the lack of certain key features or noisy data could have significantly affected the model's performance in complex scenarios.

\begin{figure}[h]
  \centering
  \includegraphics[width=\linewidth]{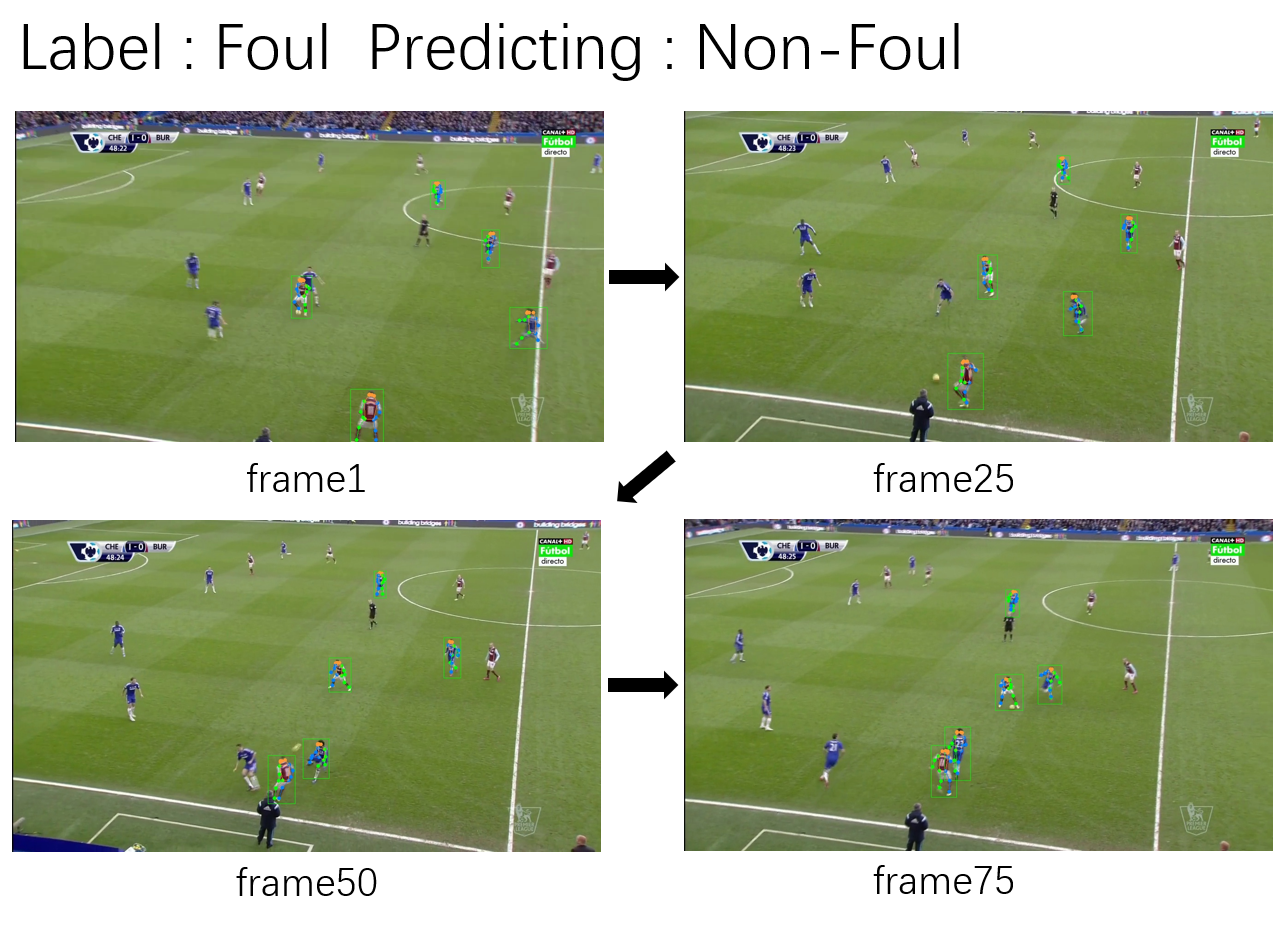}
  \caption{Qualitative evaluation using an example of missed detection in our model (FutureFoul).}
  \label{fig:miss}
\end{figure}

    For example, in Figure \ref{fig:miss}, the movement of the athletes was relatively small and the length and width of the bbox did not change significantly. In fact, foul behavior was about to appear on the two people in the middle of the frame in frame 75. There was still some distance between these two people and the pose key points of the right foot part of the one on the right were not detected correctly due to players overlapping. This could have caused the model to incorrectly assume that this situation was non-foul and failed to predict the possible foul behavior.

    \subsubsection{An examples of failures: false alarm} Another type of prediction failure occurred when the system incorrectly predicted an impending foul that did not actually occur (precision: 79.0\%). This could occur because some situations were very similar to foul behavior. The model could have incorrectly associated these with fouls.

\begin{figure}[h]
  \centering
  \includegraphics[width=\linewidth]{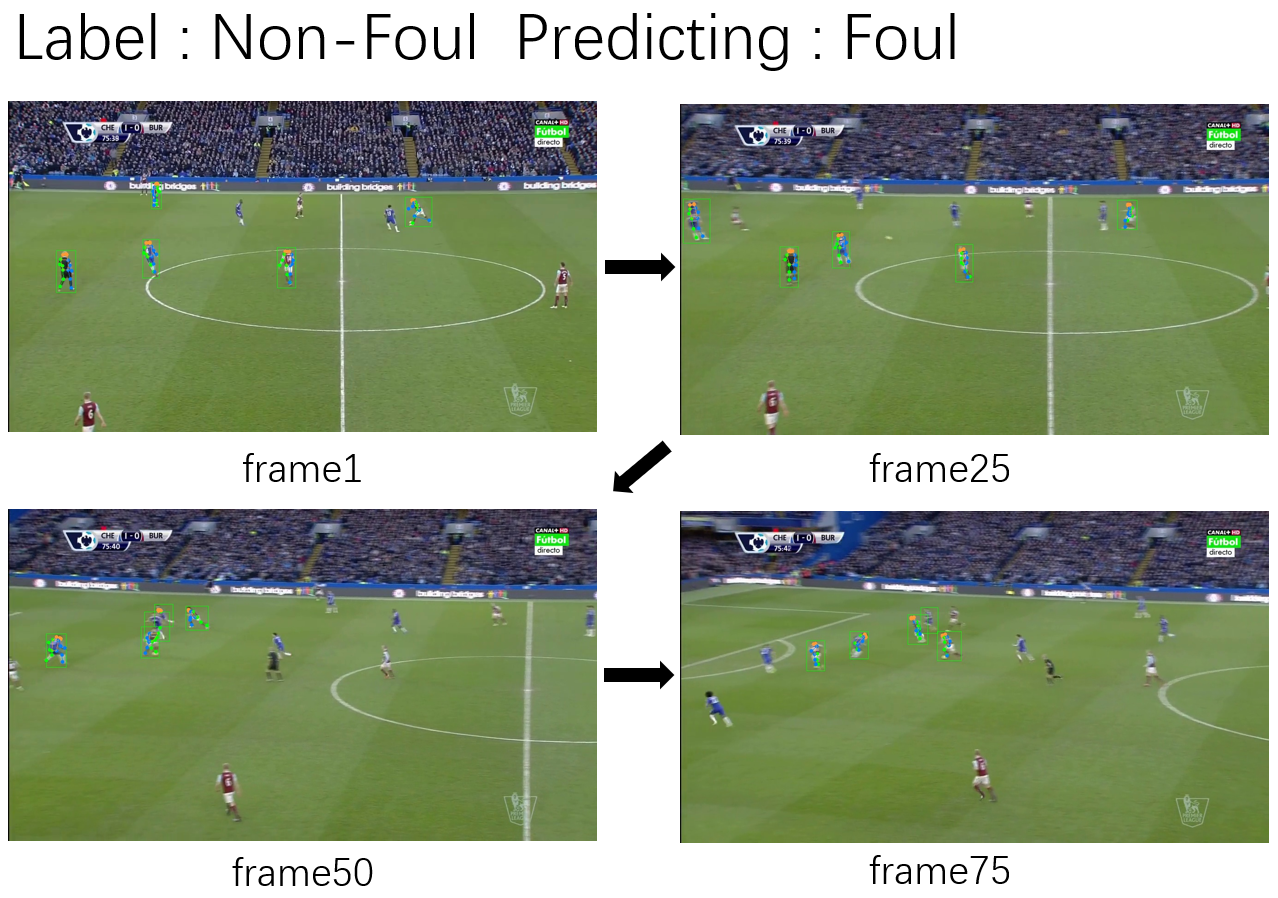}
  \caption{Qualitative evaluation using an example of false alarm. Our model warns of an upcoming foul for a normal scenario.}
  \label{fig:falsealarm}
\end{figure}

    In frames 50 and 75 in Figure \ref{fig:falsealarm}, it can be observed that the bbox had a lot of overlap. In addition, the characters tended to move in the same direction. However, there were more missing pose key points in some characters. Based on these observations, the model incorrectly predicted that foul play would occur. This situation was very similar to the foul scene.

    In addition, we note that the inaccuracy of the bboxes in the input data and the loss of pose information could have negatively affected the accuracy of the predictions. The inaccuracy of the bbox could have made it difficult for the model to accurately locate the target, while the loss of pose information could have limited the model's use of key features. 

\section{Conclusion}

    This paper presents a soccer foul dataset containing pose and bbox and proposes a method combining video, bbox, and pose to predict foul events in soccer. The results show that the pose and bbox information play an auxiliary role in the prediction of foul behavior.
    However, there are some limitations to this study. Firstly, the accuracy of the soccer foul-related dataset itself can be improved. During target detection and tracking, target loss may affect the final prediction results. In addition, detecting pose data is also challenging in pose detection due to the potential overlap between poses.
    % In addition, learning bbox and pose data is also difficult. 
    There may be some confounding factors in the current dataset that affect the learning process of bbox and pose data.

    Despite these limitations, this study provides an important exploration and contribution to the understanding of soccer foul play. Future work could further improve the accuracy and quality of the dataset, improve algorithms for target detection, tracking, and pose detection, and explore the possibility of other combinations of data sources and models to improve the accuracy and reliability of foul prediction. This will help to provide further insight into soccer foul play and provide useful support for refereeing decisions and training in soccer matches.

\section*{Acknowledgments}
This work was financially supported by JSPS Grant Number 20H04075, 23H03282 and JST PRESTO Grant Number JPMJPR20CA.

\section*{Declarations}
\subsection*{Conflict of Interest}
The authors declare that they have no conflict of interest.
\subsection*{Data availability statements}
The datasets generated during and/or analyzed during the current study will be available in the GitHub repository. % at the following link:
%\url{https://github.com/}

% \subsection*{Compliance with Ethical Standards}

\bibliography{sn-bibliography}

\end{document}